%% file: preprint.tex
\documentclass{article} 
\usepackage{iclr2025,times}

\input{math_commands.tex}

\usepackage{hyperref}
\usepackage{url}
\usepackage{adjustbox}
\usepackage{booktabs}
\usepackage{multirow}
\usepackage{enumitem}

\title{Challenges of Multi-Modal Coreset Selection for Depth Prediction}

\author{Viktor Moskvoretskii \\
Skoltech, HSE University\\
\texttt{vvmoskvoretskii@gmail.com} \\
\And
Narek Alvandian \\
Independent Researcher
}

\iclrfinalcopy 
\begin{document}

\maketitle

\begin{abstract}
Coreset selection methods are effective in accelerating training and reducing memory requirements but remain largely unexplored in applied multimodal settings. We adapt a state-of-the-art (SoTA) coreset selection technique for multimodal data, focusing on the depth prediction task.
Our experiments with embedding aggregation and dimensionality reduction approaches reveal the challenges of extending unimodal algorithms to multimodal scenarios, highlighting the need for specialized methods to better capture inter-modal relationships.
\end{abstract}

\section{Introduction}

Modern deep learning systems require massive datasets demanding hundreds of gigabytes and even terabytes of storage \cite{imagenet} \cite{laion} as well as substantial computational resources for training. To address these computational challenges, researchers have developed coreset selection methods\cite{coreset1}\cite{coreset2}\cite{coreset3} — techniques for identifying the minimal subsets of training data that maintains model performance on a level of a model trained on the full dataset. 

However, many real-world applications, from medical diagnosis \cite{SALVI2024102134} to autonomous vehicles \cite{autonomous1} \cite{autonomous3} \cite{autonomous4} to Multi-Modal Foundation Models \cite{bachmann20244m21anytoanyvisionmodel}, require processing multiple modalities of data simultaneously. These multimodal scenarios not only amplify the computational demands but also introduce new challenges, as traditional coreset selection methods cannot be directly applied.
In this work, we extend SoTA coreset selection techniques to handle multimodal data, specifically investigating the adaptation of \citet{zhou2023datasetquantization}'s approach. 
Through extensive experimentation on depth prediction tasks, we demonstrate the limitations of current approaches and the need for specialized multimodal coreset selection methods for better modeling inter-modal relationships. We provide a code for reproducing experiments.\footnote{ \url{https://github.com/VityaVitalich/MultiModalCoreset}}

\section{Method}

We adapt the coreset selection method from previous studies \citep{zhou2023datasetquantization}, successfully applied to unimodal data, to the multimodal setting. The goal is to select a representative subset \( S \) that retains the diversity and informativeness of the original dataset \( D \). Let \( D = \{(\{x^m_i\}_{m=1}^M, y_i)\}_{i=1}^N \) denote a dataset with \( N \) samples and \( M \) modalities, where \( x^m_i \) represents the features of the \( i \)-th sample for the \( m \)-th modality, and \( y_i \) is the corresponding output. For the ease of notation, further $x_i$ will denote multimodal object $\{x^m_i\}_{m=1}^M$.

The objective is to select a coreset \( S \subseteq D \) of size \( |S| = M \ll N \) that minimizes the downstream task loss:
\[
S = \underset{S \subseteq D, |S| = M}{\text{argmin}} \ \mathcal{L}_{\text{downstream}}(S),
\]
where \( \mathcal{L}_{\text{downstream}}(S) \) is the task-specific loss incurred when training on \( S \).

Following previous approaches, we employ a submodular gain function~\citep{iyer2021submodular} generalized to multimodal data, denoted as \( P(x_i) \), to measure the importance of a multimodal sample \( x_i \) in maximizing the retained information. The gain for adding \( x_i \) to the current subset \( S_{i-1} \) is:
\[
P(x_i) = \sum_{p \in S_{i-1}} ||f(p) - f(x_i)||^2 - \sum_{p \in D \setminus S_{i-1}} ||f(p) - f(x_i)||^2,
\]
where \( f(x) \) is the embedding of a \textbf{multimodal} sample \( x \) in feature space, \( S_{i-1} \) is the current subset of size \( i-1 \), and \( D \setminus S_{i-1} \) represents the remaining samples.

The dataset \( D \) is divided into non-overlapping bins \( \{S_1, S_2, \ldots, S_N\} \) through recursive selection by maximizing submodular gain $x_k \gets {\text{argmax}} \ P(x)$ with ${x \in D \setminus \bigcup_{j=1}^{n-1} S_j}$

To enhance diversity, we follow previous studies~\citep{zhou2023datasetquantization} and uniformly sample from these bins, ensuring that even the most recently selected bin contributes equally to the final coreset.

\section{Experimental Procedure}

\textbf{Dataset:} We use the CLEVR dataset~\citep{johnson2016clevrdiagnosticdatasetcompositional}, where multimodal inputs consist of RGB image and semantic mask, the target is a depth map from Omnidata~\citep{eftekhar2021omnidata}.

\textbf{Model:} We employ MultiMAE backbone, with input and output adapters trained following the original paper~\citep{bachmann2022multimaemultimodalmultitaskmasked} and DPT output adapter~\citep{Ranftl_2021_ICCV}, trained for 40 epochs with batch size 128, with best checkpoint selected. Other technical details could be found in Appendix~\ref{sec:appendix_technical}.

\textbf{Coreset Selection:} We extract embeddings $f(x)$ from the MultiMAE transformer, as the DPT output adapter's feature map dimension is too large. All coresets are 20\% of the original dataset, obtained with $N=20$. We evaluate the following baselines: \textbf{Full}: Complete dataset used for training as a reference. \textbf{Random Coreset}: A random 20\% subset. \textbf{Token Aggregation}: Concatenation, mean, or sum of embeddings. \textbf{Dimensionality Reduction}: Applying PCA or UMAP~\citep{mcinnes2020umapuniformmanifoldapproximation} to concatenation of tokens before coreset selection.

\begin{table}[t!]
\centering
\caption{Percentage of quality retained relative to the Full Dataset for Validation RMSE, Validation Loss, and Training Loss, evaluated after training with coresets selected using each method.}
\label{tab:percentage_retained}
\begin{adjustbox}{max width=\textwidth}
\begin{tabular}{lccccc}
\toprule
\textbf{Method} & \textbf{Aggregation}  & \textbf{Dimension} & \textbf{Val RMSE, \%} & \textbf{Val Loss, \%} & \textbf{Train Loss, \%} \\ 
\midrule
Full Dataset & - & - & 100.00 & 100.00 & 100.00 \\

\midrule
Random Coreset  & - & - & 50.23  & 46.33  & 55.32 \\

\midrule
\multirow{3}{*}{Coreset} & Concat & 301.824 & 49.08  & 47.41  & 51.81 \\
 & Mean & 768 & 51.54  & 47.90  & 52.63 \\
  & Sum & 768 & 51.11  & 52.12  & 54.47 \\

\midrule
\multirow{4}{*}{Coreset w/ PCA} & Concat & 512 & 47.73  & 45.43  & 45.37 \\
 & Concat & 1024 & 55.93  & 50.00  & 54.55 \\
 & Concat & 2048 & 47.90  & 43.23  & 52.46 \\
 & Concat & 4096 & 44.00  & 36.50  & 51.77 \\
\midrule

\multirow{2}{*}{Coreset w/ UMAP} & Concat & 512 & 49.29  & 49.23  & 47.44 \\
 & Concat & 1024 & 50.53  & 45.27  & 48.31 \\
\bottomrule
\end{tabular}
\end{adjustbox}
\end{table}

\section{Results \& Discussion}

Our results in Table~\ref{tab:percentage_retained} show that coreset selection methods lead to a 50\% performance drop compared to the full dataset, with minimal improvement over random coreset. The best performance is achieved with PCA (1024 features), but the improvement is incremental, and UMAP shows no consistent gain.

We observe worse convergence relative to the full dataset, likely due to reduced data representativeness, making coresets behave similarly to random selection. Attempts to use a linear output adapter for bottleneck embeddings failed, confirming the necessity of the DPT adapter for depth prediction.

\section{Conclusion}

We address the challenge of multimodal coreset selection, essential for modern applications, and present an adaptation of SoTA coreset selection method to the multimodal setting. Testing on depth prediction reveals close to random selection, highlighting the need for further exploration of multimodal coreset selection techniques.

\bibliography{iclr2025}
\bibliographystyle{iclr2025}

\newpage 

\appendix
\section{Technical Details} \label{sec:appendix_technical}

Training was performed using the Adam optimizer with learning rate \( 5 \times 10^{-5} \), \(\beta_1 = 0.9\), \(\beta_2 = 0.99\), no weight decay and cosine annealing scheduler. The training was conducted on an NVIDIA A100 GPU.


\end{document}

%% file: math_commands.tex
\usepackage{amsmath,amsfonts,bm}

\def\eqref#1{equation~\ref{#1}}

\def\1{\bm{1}}

\DeclareMathAlphabet{\mathsfit}{\encodingdefault}{\sfdefault}{m}{sl}
\SetMathAlphabet{\mathsfit}{bold}{\encodingdefault}{\sfdefault}{bx}{n}













%% file: preprint.bbl
\begin{thebibliography}{17}
\providecommand{\natexlab}[1]{#1}
\providecommand{\url}[1]{\texttt{#1}}
\expandafter\ifx\csname urlstyle\endcsname\relax
  \providecommand{\doi}[1]{doi: #1}\else
  \providecommand{\doi}{doi: \begingroup \urlstyle{rm}\Url}\fi

\bibitem[Bachmann et~al.(2022)Bachmann, Mizrahi, Atanov, and Zamir]{bachmann2022multimaemultimodalmultitaskmasked}
Roman Bachmann, David Mizrahi, Andrei Atanov, and Amir Zamir.
\newblock Multimae: Multi-modal multi-task masked autoencoders, 2022.
\newblock URL \url{https://arxiv.org/abs/2204.01678}.

\bibitem[Bachmann et~al.(2024)Bachmann, Kar, Mizrahi, Garjani, Gao, Griffiths, Hu, Dehghan, and Zamir]{bachmann20244m21anytoanyvisionmodel}
Roman Bachmann, Oğuzhan~Fatih Kar, David Mizrahi, Ali Garjani, Mingfei Gao, David Griffiths, Jiaming Hu, Afshin Dehghan, and Amir Zamir.
\newblock 4m-21: An any-to-any vision model for tens of tasks and modalities, 2024.
\newblock URL \url{https://arxiv.org/abs/2406.09406}.

\bibitem[Caesar et~al.(2020)Caesar, Bankiti, Lang, Vora, Liong, Xu, Krishnan, Pan, Baldan, and Beijbom]{autonomous4}
Holger Caesar, Varun Kumar~Reddy Bankiti, Alex Lang, Sourabh Vora, Venice~Erin Liong, Qiang Xu, Anush Krishnan, Yu~Pan, Giancarlo Baldan, and Oscar Beijbom.
\newblock nuscenes: A multimodal dataset for autonomous driving.
\newblock pp.\  11618--11628, 06 2020.
\newblock \doi{10.1109/CVPR42600.2020.01164}.

\bibitem[Chen et~al.(2012)Chen, Welling, and Smola]{coreset3}
Yutian Chen, Max Welling, and Alex Smola.
\newblock Super-samples from kernel herding, 2012.
\newblock URL \url{https://arxiv.org/abs/1203.3472}.

\bibitem[Coleman et~al.(2020)Coleman, Yeh, Mussmann, Mirzasoleiman, Bailis, Liang, Leskovec, and Zaharia]{coreset2}
Cody Coleman, Christopher Yeh, Stephen Mussmann, Baharan Mirzasoleiman, Peter Bailis, Percy Liang, Jure Leskovec, and Matei Zaharia.
\newblock Selection via proxy: Efficient data selection for deep learning, 2020.
\newblock URL \url{https://arxiv.org/abs/1906.11829}.

\bibitem[Cui et~al.(2022)Cui, Chen, Chu, Chen, Tian, Li, and Cao]{autonomous1}
Yaodong Cui, Ren Chen, Wenbo Chu, Long Chen, Daxin Tian, Ying Li, and Dongpu Cao.
\newblock Deep learning for image and point cloud fusion in autonomous driving: A review.
\newblock \emph{IEEE Transactions on Intelligent Transportation Systems}, 23:\penalty0 722--739, 02 2022.
\newblock \doi{10.1109/TITS.2020.3023541}.

\bibitem[Eftekhar et~al.(2021)Eftekhar, Sax, Malik, and Zamir]{eftekhar2021omnidata}
Ainaz Eftekhar, Alexander Sax, Jitendra Malik, and Amir Zamir.
\newblock Omnidata: A scalable pipeline for making multi-task mid-level vision datasets from 3d scans.
\newblock In \emph{Proceedings of the IEEE/CVF International Conference on Computer Vision}, pp.\  10786--10796, 2021.

\bibitem[Iyer et~al.(2021{\natexlab{a}})Iyer, Khargoankar, Bilmes, and Asanani]{iyer2021submodular}
Rishabh Iyer, Ninad Khargoankar, Jeff Bilmes, and Himanshu Asanani.
\newblock Submodular combinatorial information measures with applications in machine learning.
\newblock In \emph{Algorithmic Learning Theory}, pp.\  722--754. PMLR, 2021{\natexlab{a}}.

\bibitem[Iyer et~al.(2021{\natexlab{b}})Iyer, Khargonkar, Bilmes, and Asnani]{coreset1}
Rishabh Iyer, Ninad Khargonkar, Jeff Bilmes, and Himanshu Asnani.
\newblock Submodular combinatorial information measures with applications in machine learning, 2021{\natexlab{b}}.
\newblock URL \url{https://arxiv.org/abs/2006.15412}.

\bibitem[Johnson et~al.(2016)Johnson, Hariharan, van~der Maaten, Fei-Fei, Zitnick, and Girshick]{johnson2016clevrdiagnosticdatasetcompositional}
Justin Johnson, Bharath Hariharan, Laurens van~der Maaten, Li~Fei-Fei, C.~Lawrence Zitnick, and Ross Girshick.
\newblock Clevr: A diagnostic dataset for compositional language and elementary visual reasoning, 2016.
\newblock URL \url{https://arxiv.org/abs/1612.06890}.

\bibitem[McInnes et~al.(2020)McInnes, Healy, and Melville]{mcinnes2020umapuniformmanifoldapproximation}
Leland McInnes, John Healy, and James Melville.
\newblock Umap: Uniform manifold approximation and projection for dimension reduction, 2020.
\newblock URL \url{https://arxiv.org/abs/1802.03426}.

\bibitem[Ranftl et~al.(2021)Ranftl, Bochkovskiy, and Koltun]{Ranftl_2021_ICCV}
Ren\'e Ranftl, Alexey Bochkovskiy, and Vladlen Koltun.
\newblock Vision transformers for dense prediction.
\newblock In \emph{Proceedings of the IEEE/CVF International Conference on Computer Vision (ICCV)}, pp.\  12179--12188, October 2021.

\bibitem[Russakovsky et~al.(2015)Russakovsky, Deng, Su, Krause, Satheesh, Ma, Huang, Karpathy, Khosla, Bernstein, et~al.]{imagenet}
Olga Russakovsky, Jia Deng, Hao Su, Jonathan Krause, Sanjeev Satheesh, Sean Ma, Zhiheng Huang, Andrej Karpathy, Aditya Khosla, Michael Bernstein, et~al.
\newblock Imagenet large scale visual recognition challenge.
\newblock \emph{International journal of computer vision}, 115:\penalty0 211--252, 2015.

\bibitem[Salvi et~al.(2024)Salvi, Loh, Seoni, Barua, García, Molinari, and Acharya]{SALVI2024102134}
Massimo Salvi, Hui~Wen Loh, Silvia Seoni, Prabal~Datta Barua, Salvador García, Filippo Molinari, and U.~Rajendra Acharya.
\newblock Multi-modality approaches for medical support systems: A systematic review of the last decade.
\newblock \emph{Information Fusion}, 103:\penalty0 102134, 2024.
\newblock ISSN 1566-2535.
\newblock \doi{https://doi.org/10.1016/j.inffus.2023.102134}.
\newblock URL \url{https://www.sciencedirect.com/science/article/pii/S1566253523004505}.

\bibitem[Schuhmann et~al.(2022)Schuhmann, Beaumont, Vencu, Gordon, Wightman, Cherti, Coombes, Katta, Mullis, Wortsman, et~al.]{laion}
Christoph Schuhmann, Romain Beaumont, Richard Vencu, Cade Gordon, Ross Wightman, Mehdi Cherti, Theo Coombes, Aarush Katta, Clayton Mullis, Mitchell Wortsman, et~al.
\newblock Laion-5b: An open large-scale dataset for training next generation image-text models.
\newblock \emph{Advances in Neural Information Processing Systems}, 35:\penalty0 25278--25294, 2022.

\bibitem[Yeong et~al.(2021)Yeong, Velasco-Hernandez, Barry, and Walsh]{autonomous3}
De~Jong Yeong, Gustavo Velasco-Hernandez, John Barry, and Joseph Walsh.
\newblock Sensor and sensor fusion technology in autonomous vehicles: A review.
\newblock \emph{Sensors}, 21:\penalty0 2140, 03 2021.
\newblock \doi{10.3390/s21062140}.

\bibitem[Zhou et~al.(2023)Zhou, Wang, Gu, Peng, Lian, Zhang, You, and Feng]{zhou2023datasetquantization}
Daquan Zhou, Kai Wang, Jianyang Gu, Xiangyu Peng, Dongze Lian, Yifan Zhang, Yang You, and Jiashi Feng.
\newblock Dataset quantization, 2023.
\newblock URL \url{https://arxiv.org/abs/2308.10524}.

\end{thebibliography}
